%% file: paper.tex
\documentclass[11pt, conference]{IEEEtran}
\IEEEoverridecommandlockouts
\usepackage{amsmath,amssymb,amsfonts}
\usepackage{graphicx}
\usepackage{textcomp}
\usepackage[hidelinks, unicode]{hyperref}
\usepackage[T5]{fontenc}
\usepackage[utf8]{inputenc}
\usepackage{lmodern}
\usepackage{xurl}
\usepackage{subcaption}
\usepackage{algorithm}
\usepackage{algpseudocode}
\usepackage{booktabs}
\usepackage{flushend}
\usepackage{tikz}
\usetikzlibrary{arrows.meta} 

\newcommand{\bbar}{%
  \ooalign{b\cr\kern -0.1em 
    \begin{tikzpicture}[baseline=(X.base)]
      \node[inner sep=0pt] (X) {\phantom{b}};
      \draw[line width=0.075ex, line cap=round] 
        (-0.16em, 0.45ex) -- (0.125em, 0.45ex); 
    \end{tikzpicture}\hss}%
}

\begin{document}

\pagestyle{plain}
\title{Low-Resource Heuristics for Bahnaric Optical Character Recognition Improvement}
\author{
    \IEEEauthorblockN{Phat Tran$^*$, Phuoc Pham$^*$, Hung Trinh$^*$, Tho Quan}
    \IEEEauthorblockA{Faculty of Computer Science and Engineering \\
    Ho Chi Minh City University of Technology\\
    \texttt{\{phat.tran.k19, phuoc.phamcse206, hung.trinh\_hungking, qttho\}@hcmut.edu.vn}}
    \thanks{$^*$These authors contributed equally to this work.}
}
\maketitle

\begin{abstract}
Bahnar, a minority language spoken across Vietnam, Cambodia, and Laos, faces significant preservation challenges due to limited research and data availability. This study addresses the critical need for accurate digitization of Bahnar language documents through optical character recognition (OCR) technology. Digitizing scanned paper documents poses significant challenges, as degraded image quality from broken or blurred areas introduces considerable OCR errors that compromise information retrieval systems. We propose a comprehensive approach combining advanced table and non-table detection techniques with probability-based post-processing heuristics to enhance recognition accuracy. Our method first applies detection algorithms to improve input data quality, then employs probabilistic error correction on OCR output. Experimental results indicate a substantial improvement, with recognition accuracy increasing from 72.86\% to 79.26\%. This work contributes valuable resources for Bahnar language preservation and provides a framework applicable to other minority language digitization efforts.

\end{abstract}

\begin{IEEEkeywords}
Bahnar language, OCR, low-resource language, post-OCR correction, language modeling, table detection
\end{IEEEkeywords}

\section{Introduction}
Vietnam is home to 54 officially recognized ethnic groups, with the Kinh majority comprising 85.4\% of the population, and the remaining 53 minority groups collectively accounting for just 14.6\% \cite{gso2020completed}. Among these, the Bahnar people number approximately 286,910, representing less than 0.3\% of the national population. As digital technologies increasingly shape cultural preservation and access, safeguarding the linguistic heritage of minority communities has become a pressing concern. Yet, minority languages continue to face significant barriers to digital representation, with nearly half of the world’s writing systems lacking representation in modern digital infrastructure \cite{gla4-huot}.

The Bahnar language presents unique challenges for natural language processing research. Its rich morphological structure, combined with severe data scarcity, makes tasks such as language modeling and spelling correction particularly difficult. Most existing Bahnar textual resources consist of undigitized documents, including scanned materials affected by noise, blurred text, and typographical errors. Digitizing these materials is essential for creating foundational datasets that can enable advanced natural language processing (NLP) applications, including automatic error correction, language modeling, and preservation of endangered linguistic heritage.

\subsection{Motivation}
This study aims to create comprehensive corpus repositories for the Bahnar language through systematic digitization of existing resources. Modern machine learning and deep learning techniques have achieved remarkable success in NLP tasks, but their application has been predominantly limited to high-resource languages \cite{hedderich2021survey}. Low-resource and minority languages like Bahnar require specialized approaches that can operate effectively with limited training data. Recent work has shown that optical character recognition (OCR) technology, combined with post-processing techniques, can successfully digitize documents in low-resource languages, particularly those using extended Latin scripts \cite{rijhwani2020ocr, agarwal2024concise}.

Our digitization pipeline enables the creation of machine-readable text corpora for linguistic analysis, the development of character-level language models for Bahnar, automatic error correction systems, and the preservation of cultural materials that might otherwise be lost. By integrating contemporary scientific advances with Bahnar language processing, we aim to provide tools that support language preservation and revitalization efforts for ethnic minority communities.

\subsection{Challenges}
The global trend toward linguistic homogenization threatens many minority languages with decline or extinction. Preservation and promotion of ethnic minority languages and scripts are essential to maintaining cultural identities and ensuring equal rights among ethnic groups \cite{anh2025preservingMinority}. For languages like Bahnar, which has received limited research attention, the challenge is compounded by the lack of baseline resources, trained models, and established methodologies.

Machine learning and deep learning techniques have exhibited exceptional performance in various NLP tasks, yet their success depends on large volumes of training data \cite{devlin2019bertpretrainingdeepbidirectional, vaswani2023attentionneed}. These methods have been extensively developed for high-resource languages but require significant adaptation for low-resource languages. This creates a formidable challenge for researchers and communities working to apply NLP techniques to ethnic minority languages.

Additionally, processing tabular data is crucial for Bahnar language digitization at the lexical level. Most dictionary resources are structured in tabular format, typically with keyword columns paired with definition columns. Table detection and extraction from document images remain challenging problems due to diverse table layouts, varying encoding formats, and the presence of complex structures \cite{Goilani2017, schreiber2017deepdesrt}. OCR systems frequently produce errors when processing scanned documents containing degraded, blurred, or broken text areas, negatively impacting information retrieval quality \cite{virk2021novel}.

\subsection{Discussion}
We address table extraction using traditional image processing techniques implemented with OpenCV, leveraging the presence of table separators in our dataset. Our approach exploits the geometric characteristics of horizontal and vertical table edges to identify cell boundaries through contour detection. We compute line equations for table edges and determine their intersections to locate corner coordinates for tables and individual cells. This geometric method proves effective for well-structured tables with clear separators.

For text recognition, we apply OCR processing to individual cells after table segmentation. The extracted text then undergoes post-processing using $n$-gram-based probabilistic methods to correct recognition errors. This correction stage is critical, as OCR systems typically introduce systematic errors when processing minority language documents, particularly those with limited representation in training data.

\subsection{Contribution}
Our work makes several key contributions to the field of low-resource language processing. First, we introduce a practical pipeline for applying OCR techniques and heuristic post-processing to the Bahnar language, which is a severely under-resourced minority language. Our experimental results offer a valuable demonstration for researchers working with other underdeveloped languages, especially those that are minority or endangered and face similar resource limitations.

Second, our digitization efforts yield machine-readable datasets in the Bahnar language, which represent some of the earliest resources of their kind. These datasets enable a range of downstream NLP applications, such as language modeling, machine translation, and linguistic analysis. By sharing these resources with the research community, we aim to foster further work on Bahnar and other languages with limited computational resources.

Finally, our work confirms the effectiveness of combining traditional computer vision techniques for document structure analysis with modern probabilistic methods for error correction. This hybrid approach offers a practical framework for document digitization projects in resource-constrained settings, where state-of-the-art deep learning models may be impractical due to limited training data.

\section{Related Works}
The digitization of historical and low-resource language documents through OCR has received increasing attention in recent years, with notable progress in both error detection and correction methodologies. This section reviews relevant work in three key areas: post-OCR error correction, table detection and extraction, and OCR applications for low-resource languages.
\subsection{Post-OCR Error Correction}

OCR post-processing has been extensively studied through competitions organized by the International Conference on Document Analysis and Recognition (ICDAR) in 2017 and 2019 \cite{icdar2017post, icdar2019post}. These competitions focused on two primary tasks: error detection and error correction, using evaluation datasets for English and French. The majority of participating teams employed machine learning and deep learning approaches, including attention-based neural machine translation \cite{dong2018multi}, sequence-to-sequence models, and transformer-based architectures \cite{lyu2021neural, mokhtar2018ocr}, achieving remarkable performance on high-resource languages.

However, the success of these data-intensive approaches depends critically on the availability of large training corpora, making them impractical for low-resource languages like Bahnar. Recent surveys on OCR for low-resource languages highlight that nearly half of the world's writing systems lack adequate digital representation and training resources \cite{agarwal2024concise}. For endangered and minority languages, post-OCR correction must rely on alternative strategies that operate effectively with limited data.

Simpler dictionary-based and statistical approaches have delivered encouraging results in ICDAR competitions, reducing character error rates by 13\% for English monographs and 23\% for French monographs, even with relatively low initial CERs of 1--4\% \cite{icdar2017post}. Traditional statistical and rule-based spelling correction methods have consistently improved word accuracy from 80\% to 90\%, producing single-digit word error rates (WER) \cite{tong1996statistical, bassil2012ocr, thompson2015customised}. These results have inspired our hybrid approach, combining dictionary-based lookup with statistical language models and heuristic post-processing rules.

Recent work on endangered language OCR has revealed that probabilistic methods, including $n$-gram language models and edit distance algorithms, effectively correct OCR errors without requiring extensive parallel training data. Character-level language models have proven particularly effective for morphologically rich languages with limited resources. Additionally, context-aware correction using language-specific dictionaries and morphological analyzers significantly improves accuracy for low-resource scenarios.

\subsection{Table Detection and Extraction}

Automatic extraction of tabular information from document images presents unique challenges due to the diversity of table layouts, structures, and encoding formats. Tables are crucial for processing Bahnar lexical resources, as most dictionary data is organized in tabular format with keyword columns paired with definition columns \cite{yil2018bahnar}.

Early approaches to table detection relied on heuristics and structural metadata analysis. The TINTIN system \cite{pyreddy1997tintin} utilized structural data to identify tables and their constituent fields. Kasar et al. \cite{kasar2013learning} employed Support Vector Machine (SVM) classifiers to distinguish table regions by detecting crossing horizontal and vertical lines combined with low-level image features. These traditional methods proved effective for well-structured documents but struggled with complex or degraded table layouts.

The emergence of deep learning techniques revolutionized table detection capabilities. Gilani et al. \cite{Goilani2017} pioneered the application of deep learning to table detection using a Faster R-CNN-based architecture, introducing distance-based data augmentation to enhance model accuracy. Subsequent work extended these approaches with more sophisticated architectures. Schreiber et al. \cite{schreiber2017deepdesrt} proposed DeepDeSRT, combining detection and structure recognition in a unified framework that handles both table localization and cell-level structure analysis. 

For our implementation, we adopt a hybrid approach leveraging traditional computer vision techniques implemented with OpenCV \cite{opencv_library}. Given that tables in our Bahnar dataset consistently contain explicit separators (borders), we exploit geometric properties of horizontal and vertical edges through contour detection and line intersection computation. This approach offers computational efficiency and interpretability while achieving satisfactory accuracy for our structured documents. The extracted table cells are then processed individually through OCR, with subsequent post-processing to correct recognition errors.

\subsection{OCR for Low-Resource Languages}

The application of OCR technology to low-resource and minority languages presents distinct challenges compared to well-studied languages like English or Chinese. Limited availability of training data, lack of standardized fonts, and absence of robust language models constrain the effectiveness of modern deep learning approaches.

Recent studies have introduced effective techniques for adapting OCR systems to low-resource settings. Rijhwani et al. \cite{rijhwani2020ocr} found that post-OCR correction using character-level language models and phonetic edit distances can improve accuracy for endangered languages without requiring large parallel corpora. Their approach leverages linguistic properties and small dictionaries to constrain the search space for corrections, achieving marked error reductions with minimal resources.

Transfer learning from high-resource to low-resource languages has shown promise, particularly for scripts sharing similar visual characteristics. However, for unique scripts or those with limited representation in existing models, simpler probabilistic approaches often prove more practical and effective. Character-level modeling, which requires less data than word-level approaches, has emerged as a particularly suitable strategy for morphologically complex minority languages.

For Vietnamese minority languages, including Bahnar, previous work has explored basic language modeling and spelling correction \cite{linh2021building}. However, comprehensive digitization pipelines integrating document structure analysis, OCR, and post-processing remain limited. Our work builds upon these foundations, introducing a complete pipeline suitable for digitizing Bahnar language resources from scanned documents.
\section{Data preprocessing}
\subsection{Image Cropping}
Our dataset consists of scanned pages from the Bahnar Dialect Dictionary \cite{yil2018bahnar}. Pages were scanned using a smartphone camera with a limited depth of field, which resulted in images containing extraneous artifacts and regions beyond the dictionary pages. These artifacts include background elements and peripheral content that reduce the quality of the data for subsequent processing steps.

\begin{figure}[!htbp]
    \centering
    \begin{subfigure}[b]{0.49\linewidth}
        \centering
        \includegraphics[width=\linewidth]{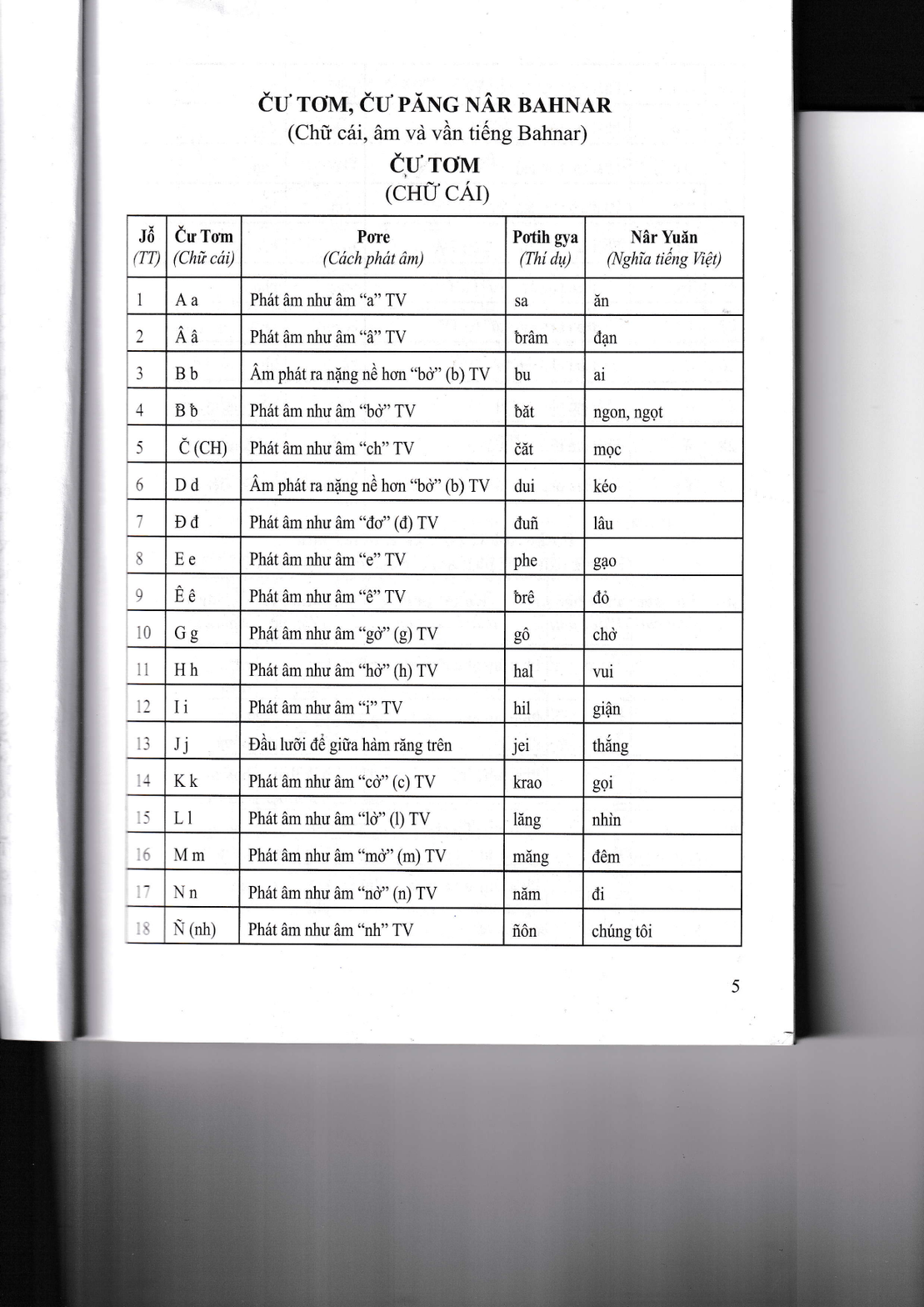}
        \caption{Page 5.}
        \label{fig:bahnar_page1}
    \end{subfigure}
    \hfill
    \begin{subfigure}[b]{0.49\linewidth}
        \centering
        \includegraphics[width=\linewidth]{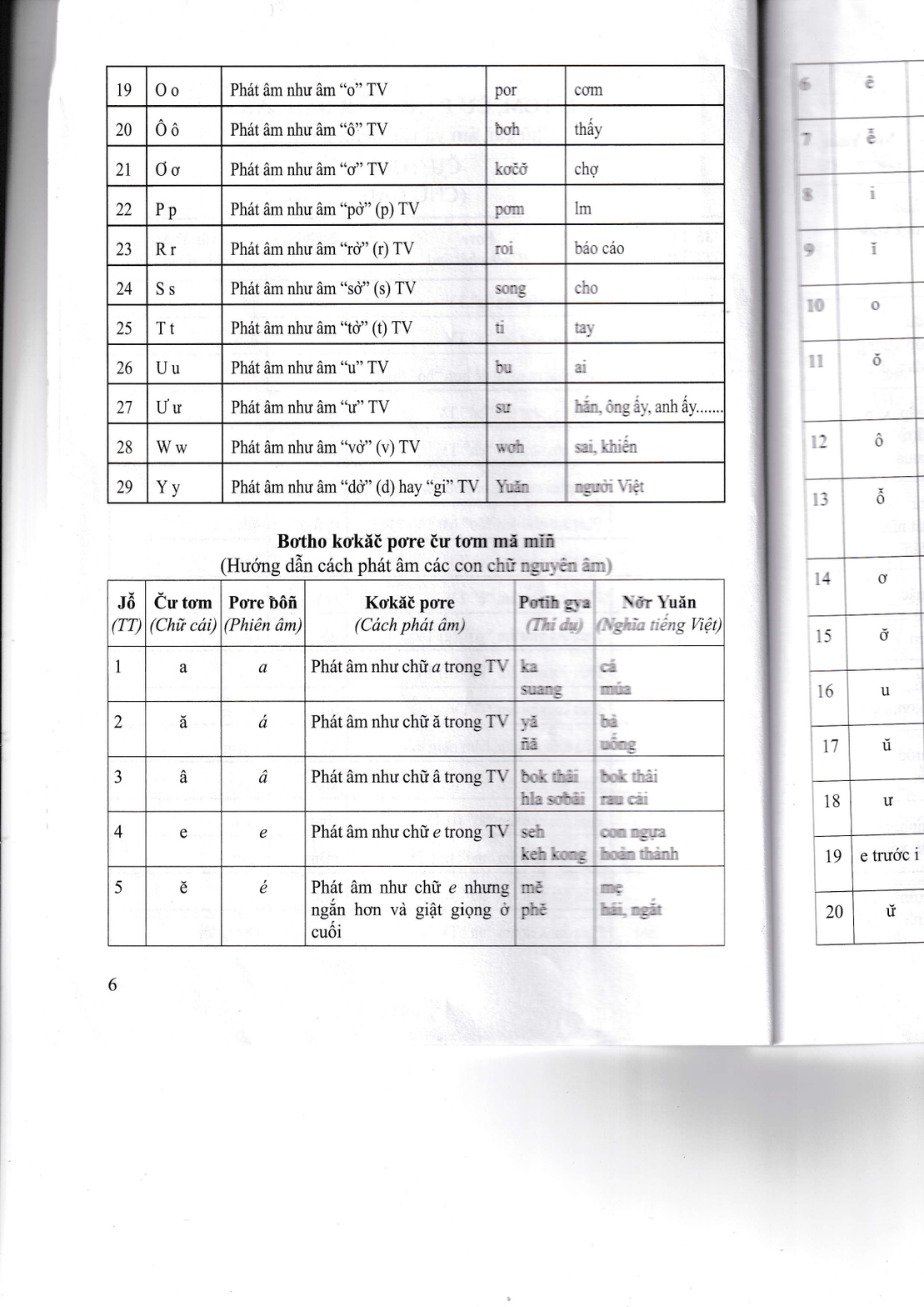}
        \caption{Page 6.}
        \label{fig:bahnar_page2}
    \end{subfigure}
    \caption{Original images of Bahnar Dialect Dictionary.}
    \label{fig:Bahnar_dialect_dict}
\end{figure}

\begin{figure}[htbp]
    \centering
    \begin{subfigure}[b]{0.49\linewidth}
        \centering
        \includegraphics[width=\linewidth]{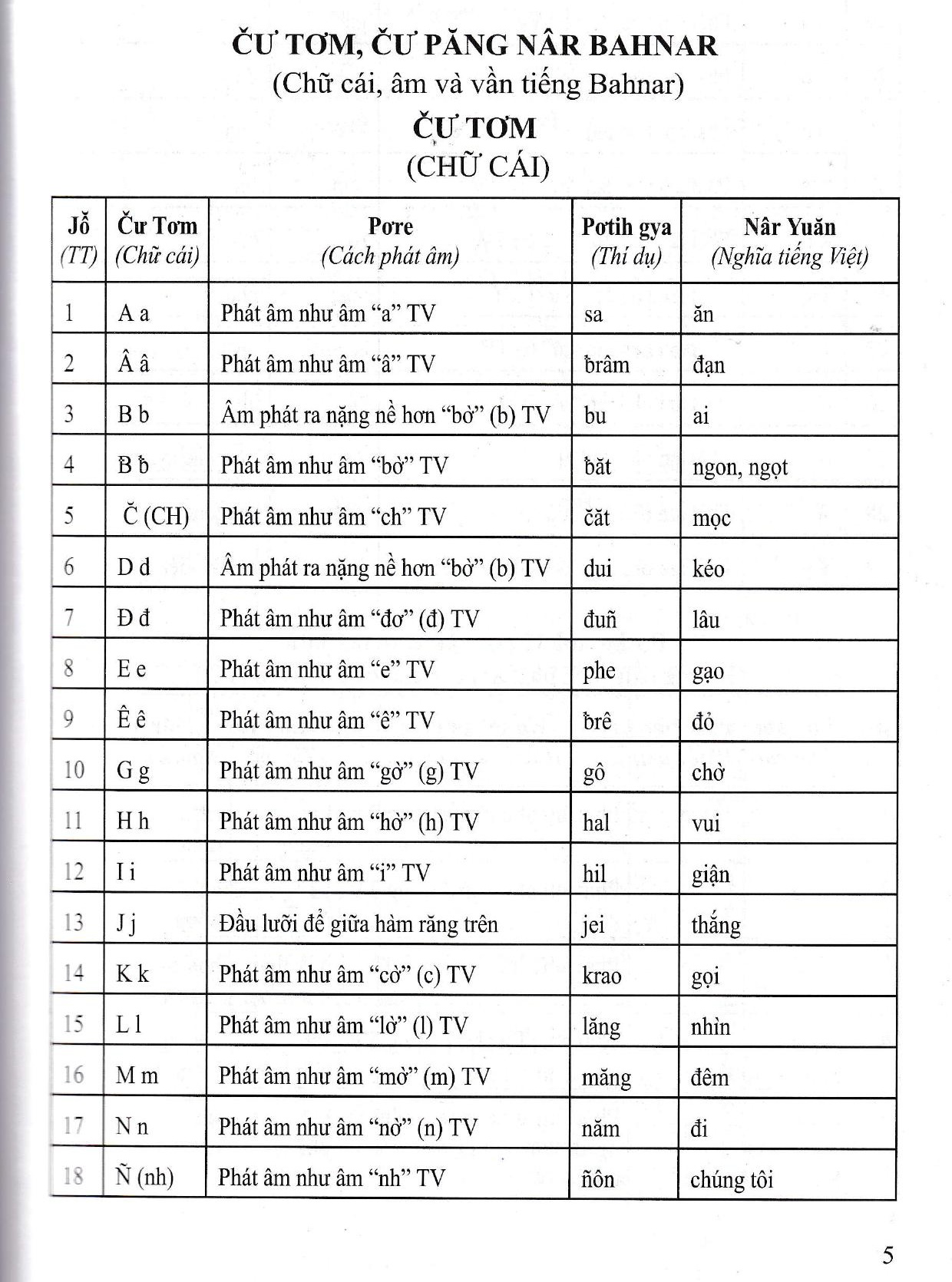}
        \caption{Cropped page 5.}
        \label{fig:bahnar_cropped1}
    \end{subfigure}
    \hfill
    \begin{subfigure}[b]{0.49\linewidth}
        \centering
        \includegraphics[width=\linewidth]{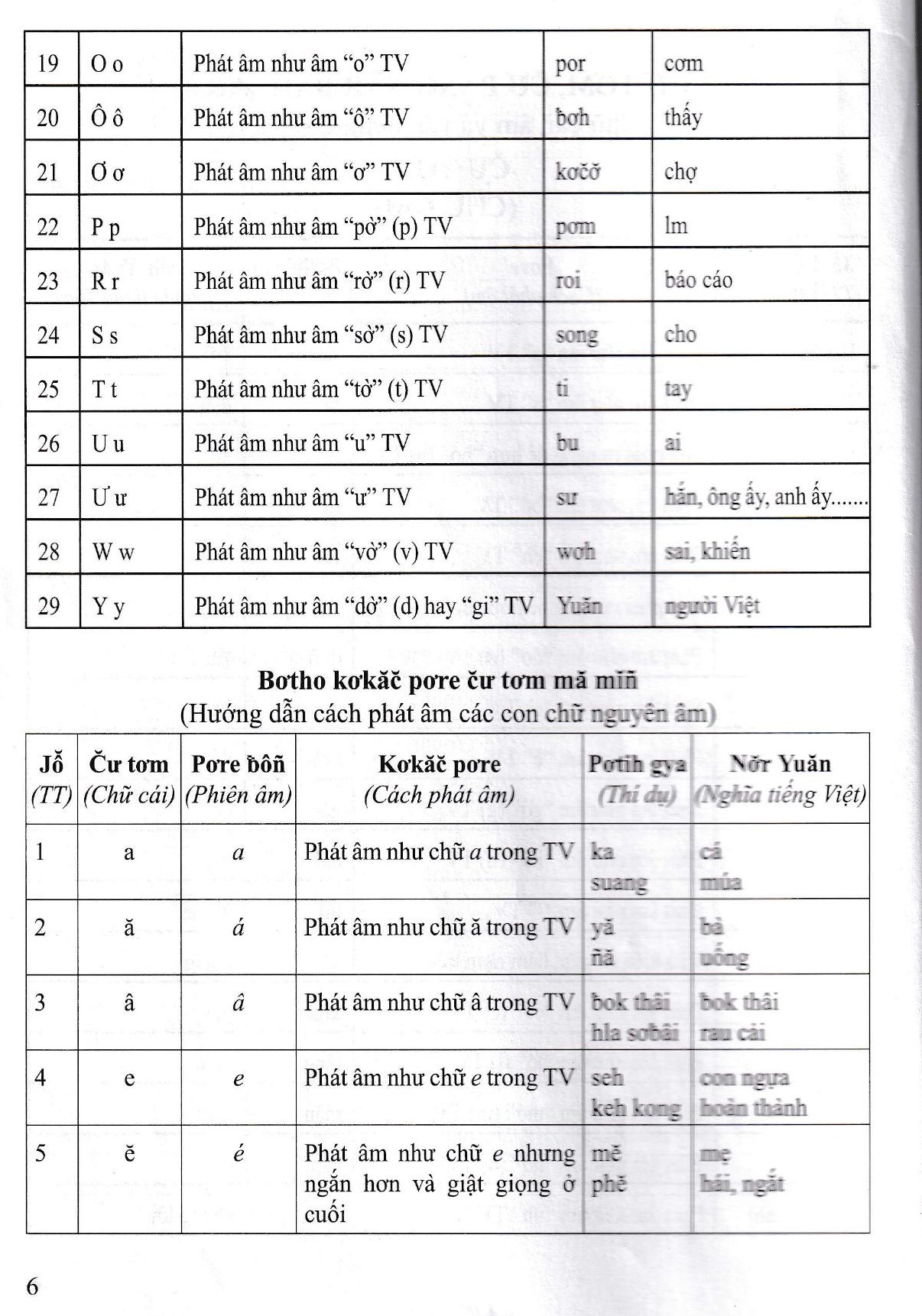}
        \caption{Cropped page 6.}
        \label{fig:bahnar_cropped2}
    \end{subfigure}
    \caption{Cropped images of Bahnar Dialect Dictionary.}
    \label{fig:Bahnar_dialect_dict_cropped}
\end{figure}

To mitigate these issues, we implement a preprocessing pipeline that crops images to isolate the relevant dictionary content before applying OCR. Figure \ref{fig:Bahnar_dialect_dict} presents three representative samples of the original scanned images, while Figure \ref{fig:Bahnar_dialect_dict_cropped} displays the corresponding cropped outputs.

\subsection{Binary Image Transforming}
Following the cropping stage, we convert each image to binary form through adaptive thresholding. Traditional global thresholding methods rely on manually selected threshold values to classify each pixel as either foreground (1) or background (0) based on its intensity. To avoid this subjective parameter selection and ensure reproducibility, we employ Otsu's method \cite{Yousefi2015}, which automatically determines the optimal threshold value by analyzing the image histogram.

Otsu's algorithm operates on the principle that document images typically exhibit bimodal intensity distributions, with distinct peaks corresponding to text and background regions. The method computes a threshold value \(t\) that maximizes the between-class variance, which measures the separability between background and foreground classes, expressed as:

\[
\sigma_b^2(t) = \omega_{bg}(t)\omega_{fg}(t)[\mu_{bg}(t) - \mu_{fg}(t)]^2
\]

where \(\omega_{bg}(t)\) and \(\omega_{fg}(t)\) represent the probability distributions of background and foreground pixels at threshold \(t\), respectively, and \(\mu_{bg}(t)\) and \(\mu_{fg}(t)\) denote the mean intensity values of each class.

The probability weights are computed from the normalized histogram as follows:

\[
\omega_{bg}(t) = \sum_{i=0}^{t} p(i)
\]

\[
\omega_{fg}(t) = \sum_{i=t+1}^{L-1} p(i) = 1 - \omega_{bg}(t)
\]

where \(p(i) = \frac{n_i}{N}\) represents the normalized histogram probability at intensity level \(i\), \(n_i\) is the count of pixels with intensity \(i\), \(N\) is the total pixel count, and \(L\) is the number of intensity levels (typically 256 for 8-bit grayscale images).

The mean intensity values for each class are calculated using:

\[
\mu_{bg}(t) = \frac{\sum_{i=0}^{t} i \cdot p(i)}{\omega_{bg}(t)}
\]

\[
\mu_{fg}(t) = \frac{\sum_{i=t+1}^{L-1} i \cdot p(i)}{\omega_{fg}(t)}
\]

The algorithm iteratively evaluates potential threshold values to identify the one that maximizes \(\sigma_b^2(t)\), thereby achieving optimal separability between background and foreground regions.

\subsection{Edge Detection}

Following binary transformation, we extract horizontal and vertical edges from the image using morphological operations. We employ the \texttt{getStructuringElemen} function to define structural elements for edge detection, followed by erosion and dilation operations to refine and strengthen the detected edge segments.

\subsubsection{Line Detection}

Our line detection approach is based on Hough Line Transform method \cite{duda1972hough}. We apply the \texttt{HoughLinesP} function from OpenCV to identify line segments in the preprocessed binary image. Since the initial detection often produces fragmented line segments, we implement a grouping algorithm that consolidates collinear segments into continuous main edges. This consolidation process merges nearby parallel segments that belong to the same table border, resulting in a more robust set of edge lines for subsequent processing.

\subsection{Image Rotation}

Scanned images frequently exhibit angular distortion that adversely affects line and cell detection accuracy. To correct this distortion, we straighten the image by aligning it with the table's vertical edges. Our analysis of the dataset reveals that all vertical table edges are parallel, enabling us to use any single vertical edge as a reference for rotation correction.

Consider a vertical edge defined by two endpoints with coordinates \((x_1, y_1)\) and \((x_2, y_2)\). The angle \(\theta\) between this edge and the y-axis is calculated as:

\[
\theta = \arctan\left(\frac{x_1 - x_2}{y_1 - y_2}\right)
\]

We then apply a rotation transformation of angle \(-\theta\) using OpenCV's rotation functions to align the image with the coordinate axes, effectively removing the skew.

\subsection{Table Cell Detection}

After image rectification, we perform a second edge detection pass to obtain updated horizontal and vertical line segments. The cell boundaries are then determined by computing the intersection points of these orthogonal lines and analyzing their spatial relationships.

\subsubsection{Intersection Point Computation}

To identify cell corners, we compute intersection points between horizontal and vertical edges using a parametric line intersection algorithm. Consider two line segments: \(A\) defined by endpoints \(\mathbf{a}_1\) and \(\mathbf{a}_2\), and \(B\) defined by endpoints \(\mathbf{b}_1\) and \(\mathbf{b}_2\).

The direction vector of line \(A\) is:
\[
\mathbf{n}_a = \mathbf{a}_2 - \mathbf{a}_1 = (x_1, y_1)
\]

The normal vector of line \(A\) is:
\[
\mathbf{u}_a = (-y_1, x_1)
\]

Similarly, the direction vector of line \(B\) is:
\[
\mathbf{n}_b = \mathbf{b}_2 - \mathbf{b}_1
\]

The direction vector from \(\mathbf{a}_1\) to \(\mathbf{b}_1\) is:
\[
\mathbf{n}_p = \mathbf{a}_1 - \mathbf{b}_1
\]

The intersection point is computed using:
\[
\mathbf{P}_{\text{intersection}} = \left(\frac{\mathbf{u}_a \cdot \mathbf{n}_p}{\mathbf{u}_a \cdot \mathbf{n}_b}\right) \mathbf{n}_b + \mathbf{b}_1
\]

This parametric approach offers two key advantages: it automatically handles cases where edges are partially broken or degraded by interpolating through gaps, and it guarantees comprehensive cell detection with minimal false negatives.

\subsubsection{Cell Corner Identification}

After computing all line intersections, we obtain a set of candidate corner points. To identify the four corners of each cell, we apply a point-spreading algorithm. For each point $(x_0, y_0)$ in the intersection set, we search for the nearest intersection point to the right, $(x_1, y_0)$, and the nearest intersection point below, $(x_0, y_1)$.

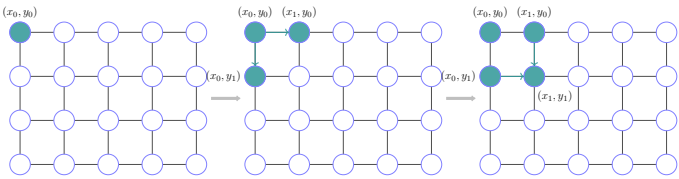
\begin{figure}[!htbp] 
    \centering   
    \resizebox{0.5\textwidth}{!}{\input{Tikz/spread_points}}
    \caption{Cell corner identification via point-spreading over line intersections.}    \label{fig:cell_detection} 
\end{figure}

If the diagonal point \((x_1, y_1)\) exists in the intersection set, then the four points \((x_0, y_0)\), \((x_1, y_0)\), \((x_0, y_1)\), and \((x_1, y_1)\) form a valid cell. Otherwise, we proceed to the next candidate point. This systematic approach ensures cells are automatically identified and ordered from top to bottom and left to right, as illustrated in Figure \ref{fig:cell_detection}.

\subsubsection{Non-Table Region Detection}

In our dataset, tables consistently appear as standalone elements spanning the full width of the page, with no text content appearing to their left or right. To separate table and non-table regions, we identify the top and bottom horizontal boundaries of each table. Using OpenCV functions, we mask the region between these boundaries to isolate the table content. The remaining regions above and below the table are extracted as separate images, which serve as inputs for the OCR model to process the non-tabular dictionary content.

\section{Heuristics for Post-processing}

To correct common errors in OCR-extracted text, we implement a post-processing heuristic that relies on a custom-built Bahnar reference dictionary. This process is divided into two phases: first, the construction of the reference dictionary from a validated vocabulary, and second, the application of a sliding-window correction algorithm that uses this dictionary.

\subsection{Reference Dictionary Construction}

We first construct a Bahnar vocabulary dataset that is accurate, well-structured, and linguistically consistent. During preprocessing, punctuation marks (periods, commas, dashes, semicolons, colons, parentheses, and quotation marks) are removed and replaced with whitespace. Multi-word phrases are tokenized into individual words.

Single-character tokens are generally filtered out. However, some exceptions are made for Bahnar diacritics: single-character words containing accents such as ` $\breve{}$ ' are retained, as the base character and diacritic are treated as distinct components. For instance, `$\breve{\text{c}}$' is counted as a two-character token. Similarly, words with multiple diacritics, such as `$\hat{\breve{\text{e}}}$', `$\hat{\breve{\text{o}}}$', `$\hat{\breve{\text{E}}}$', or `$\hat{\breve{\text{O}}}$', are counted as three-character tokens. After filtering, the validated terms are appended to a vocabulary stack.

\begin{algorithm}[!htbp]
\caption{Preprocessing and tokenization of Bahnar vocabulary entries.}
\label{algo:entry}
\begin{algorithmic}[1]
\Require Bahnar vocabulary entry
\Ensure Stack $S$ of validated single words

\State Remove special characters [\texttt{,\_-""();:.}] and replace with whitespace
\If{input contains multiple words}
    \State Tokenize into individual words
\EndIf
\For{each token $t$}
    \If{$\text{len}(t) = 1$ \textbf{and} $t$ has no diacritics}
        \State Discard $t$ \Comment{Filter out single non-diacritic characters}
    \EndIf
\EndFor

\State \Return $S$
\end{algorithmic}
\end{algorithm}

Subsequently, we construct a Bahnar dictionary from the vocabulary stack by extracting character $n$-grams. Each word is decomposed into overlapping character clusters ($n$-grams) with lengths 2--4. These clusters serve as dictionary keys, mapping to a frequency distribution of word lengths. Specifically, for an $n$-gram $c$, the dictionary $D[c]$ stores a map of \texttt{(word\_length: count)}, indicating how many times $c$ has appeared in a validated word of specific length. This structure enables frequency tracking and supports the error correction algorithm. The workflows are illustrated in Algorithm \ref{algo:entry} and Algorithm \ref{algo:stack}.

\begin{algorithm}[!htbp]
\caption{Construction of $n$-gram frequency dictionary from validated vocabulary.}
\label{algo:stack}
\begin{algorithmic}[1]
\Require Stack $S$ of validated words
\Ensure Bahnar dictionary $D$ mapping character clusters to word length frequency counts

\While{$S$ is not empty}
    \State $word \gets$ Pop from $S$
    \State $clusters \gets$ Extract $n$-grams of length $\{2, 3, 4\}$ from $word$
    \State $n \gets \text{len}(word)$
    
    \For{each cluster $c$ in $clusters$}
        \If{$D[c]$ does not exist}
            \State $D[c] \gets \text{new Map()}$ \Comment{Initialize cluster entry}
        \EndIf
        \If{$D[c][n]$ does not exist}
            \State $D[c][n] \gets 0$ \Comment{Initialize length counter}
        \EndIf
        \State $D[c][n] \gets D[c][n] + 1$ \Comment{Increment frequency count}
    \EndFor
\EndWhile

\State \Return $D$
\end{algorithmic}
\end{algorithm}

\subsection{Error Correction Heuristic}

The post-processing heuristic, described in Algorithm \ref{alg:heuristic}, corrects errors in a single-word string \texttt{str} extracted from OCR. Before heuristic correction, we apply general-case corrections for unambiguous, non-Bahnar characters, such as mapping `{]}' to `{l}', or mapping `$\ddot{\text{O}}$' and `$\check{\text{s}}$' (which do not exist in the Bahnar alphabet) to `$\breve{\text{C}}$' and `$\breve{\text{c}}$', respectively.

The algorithm uses two nested loops. The outer loop (index $i$) slides a window across the string. The inner loop (index $j$) checks for invalid $n$-grams at position $i$, prioritizing longer $n$-grams first (i.e., $j=4$, then $j=3$, then $j=2$).

To validate an $n$-gram $\texttt{tmp} = \texttt{str}[i:i+j]$, we query our reference dictionary. We define a frequency function, $\text{prob}(\texttt{tmp}, n)$, which returns the count $D[\texttt{tmp}][n]$, where $n$ is the length of the original word \texttt{str}. A low count (e.g., fewer than a threshold $\texttt{thres}=5$) suggests $\texttt{tmp}$ is an unlikely $n$-gram for a word of length $n$ and is flagged as an error.

The first invalid $n$-gram found (prioritizing length) is passed to a correction function, described in Algorithm \ref{alg:correct}, which attempts to find a valid replacement. The string is updated with the corrected subword, and the inner loop is terminated to proceed to the next position $i$.

\begin{algorithm}[!htbp]
\caption{Sliding window error detection and correction for OCR output.}
\label{alg:heuristic}
\begin{algorithmic}[1]
\Require Single word input \texttt{str} from OCR image 
\Ensure Corrected string
\State Correct general cases of \texttt{str}
\State $i \gets 0$
\State $n \gets \text{len}(\texttt{str})$
\State $\texttt{thres} \gets 5$

\While{$i < n$}
    \State $j \gets 4$
    
    \While{$i + j \leq n$ \textbf{ and } $j > 1$}
        \State $\texttt{tmp} \gets \texttt{str}[i:i+j]$
        
        \If{$\text{prob}(\texttt{tmp}, n) < \texttt{thres}$}
            \State $\texttt{tmp} \gets \text{correct}(\texttt{tmp}, n)$ \Comment{Attempt correction}
            \State $\texttt{str}[i:i+j] \gets \texttt{tmp}$ \Comment{Update string}
            \State \textbf{break} \Comment{Prioritize longest corrected $n$-gram}
        \Else
            \State $j \gets j - 1$ \Comment{$n$-gram is valid, check shorter one}
        \EndIf
    \EndWhile
    
    \State $i \gets i + 1$
\EndWhile

\State \Return \texttt{str}
\end{algorithmic}
\end{algorithm}

The \texttt{correct()} helper function, detailed in Algorithm \ref{alg:correct}, searches for a replacement for the invalid subword. It performs this search by testing every possible single-character substitution within the subword, using each character from the Bahnar alphabet.

For each potential substitution, it calculates its frequency count $\texttt{prob}(\texttt{new\_sub}, n)$. The algorithm selects the substitution that results in the subword with the \textit{highest} frequency count. If no substitution (including the original subword) achieves a count greater than or equal to the threshold, it is ``unsuitable'', and the original, uncorrected subword is returned to avoid erroneous changes.

\begin{algorithm}[!htbp]
\caption{Single-character substitution correction using frequency-based selection.}
\label{alg:correct}
\begin{algorithmic}[1]
\Require Substring input $\texttt{subword}$ and original length $n$
\Ensure Corrected result string (\texttt{res})

\State $\texttt{thres} \gets 5$
\State $\texttt{Bahnar\_ABC} \gets [\,]$ \Comment{The set of all Bahnar characters}
\State $\texttt{best\_sub} \gets \texttt{subword}$
\State $\texttt{max\_prob} \gets \texttt{prob}(\texttt{subword}, n)$

\State $i \gets 0$
\While{$i < \text{len}(\texttt{subword})$}
    \State $j \gets 0$
    \While{$j < \text{len}(\texttt{Bahnar\_ABC})$}
        \State $\texttt{new\_sub} \gets \texttt{subword}$
        \State $\texttt{new\_sub}[i] \gets \texttt{Bahnar\_ABC}[j]$
        \State $\texttt{current\_prob} \gets \texttt{prob}(\texttt{new\_sub}, n)$
        
        \If{$\texttt{current\_prob} > \texttt{max\_prob}$}
            \State $\texttt{max\_prob} \gets \texttt{current\_prob}$
            \State $\texttt{best\_sub} \gets \texttt{new\_sub}$
        \EndIf
        
        \State $j \gets j + 1$
    \EndWhile
    \State $i \gets i + 1$
\EndWhile

\If{$\texttt{max\_prob} < \texttt{thres}$}
    \State \Return $\texttt{subword}$ \Comment{No suitable replacement found, return original}
\Else
    \State \Return $\texttt{best\_sub}$ \Comment{Return substitution with highest frequency}
\EndIf
\end{algorithmic}
\end{algorithm}

\section{Methodology}
\subsection{Dataset}
The Bahnar Dialect Dictionary, published by the Gia Lai Department of Education and Training, serves as the primary resource for this study \cite{yil2018bahnar}. The corpus comprises 115 images containing heterogeneous la ts, including both tabular and non-tabular structures, with extensive Bahnar-Vietnamese bilingual vocabulary.

The Bahnar orthography presents challenges for conventional OCR systems due to its extensive use of special characters, including `$\breve{\text{c}}$', `$\breve{\text{e}}$', `$\hat{\breve{\text{e}}}$', `$\breve{\text{i}}$', `$\tilde{\text{n}}$', `$\breve{\text{o}}$', `$\hat{\breve{\text{o}}}$', `$\breve{\text{ơ}}$', `$\breve{\text{u}}$', and `$\breve{\text{ư}}$', among others. These diacritical marks, uncommon in high-resource languages, necessitate specialized post-processing strategies. Furthermore, the presence of noise artifacts and scanning anomalies in the digitized images requires robust preprocessing techniques to ensure recognition quality.

\subsection{Tesseract}
We employed Tesseract, an open-source OCR engine renowned for its reliability and extensibility in document digitization tasks \cite{Smith2007}. Tesseract functions as the core text extraction module in our pipeline, processing both tabular and non-tabular regions. However, Tesseract's pre-trained models provide native support exclusively for high-resource languages such as English, French, Vietnamese, and Bulgarian, lacking direct support for low-resource minority languages like Bahnar. This limitation motivates our heuristic-based correction approach to address character-level recognition errors.

Tesseract provides 14 Page Segmentation Modes (PSMs) that define how the engine interprets document layout. Through systematic evaluation across all PSM configurations, we identified PSM 6 as optimal for our dataset, as it assumes a single uniform block of text, an assumption consistent with the dictionary's structured layout and uniform typography. Our final configuration utilizes \texttt{-l vie+en --oem 1 --psm 6}, leveraging Vietnamese and English language models with the LSTM-based OCR engine mode.

\subsection{Experiment with Language Modeling}
To evaluate potential improvements through language modeling, we conducted experiments using a character-level language model specifically developed for Bahnar, incorporating both left-to-right (L2R) and right-to-left (R2L) character models \cite{linh2021building}. The model was applied as a post-processing step to refine raw OCR outputs.

\begin{figure}[!htbp]
    \centering
    \begin{subfigure}[b]{\linewidth}
        \centering
        \includegraphics[width=\linewidth]{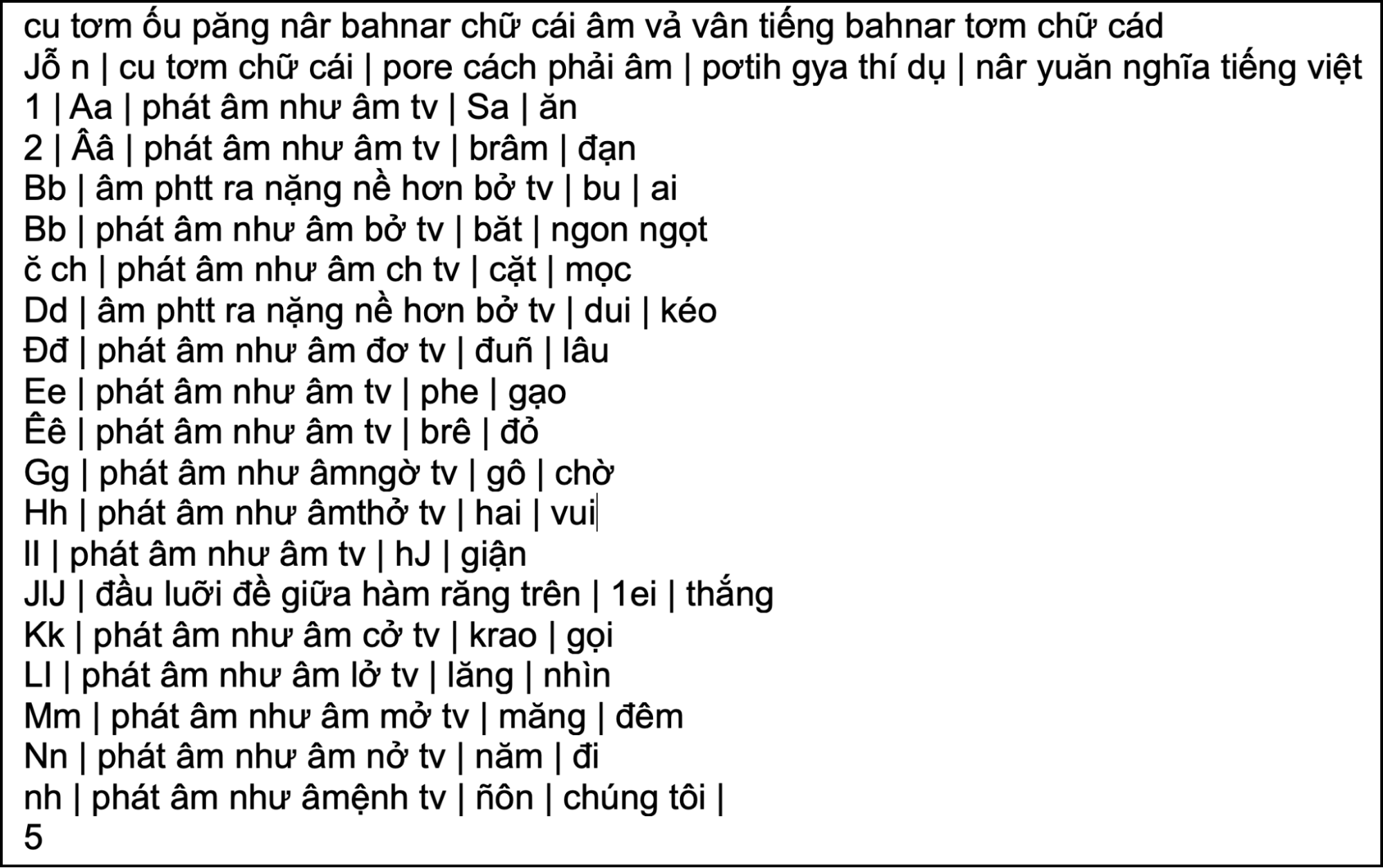}
    \end{subfigure}
    \begin{subfigure}[b]{\linewidth}
        \centering
        \includegraphics[width=\linewidth]{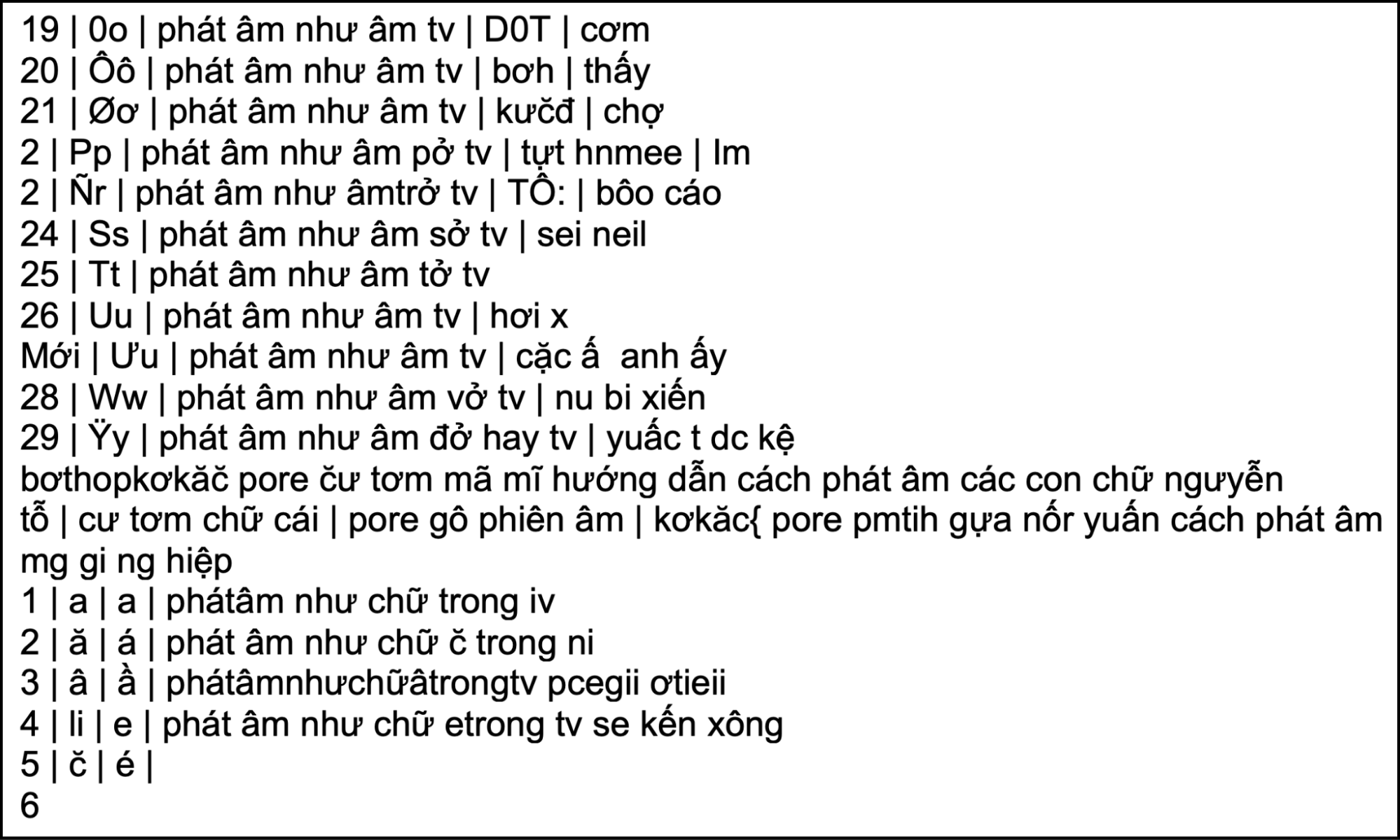}
    \end{subfigure}
    \caption{Language model-enhanced OCR output for pages 5 and 6.}
    \label{fig:ocr_lm_pages56}
\end{figure}

However, experimental results revealed suboptimal performance, which we attribute primarily to insufficient training data, a common challenge in low-resource language processing. The model frequently failed to correct genuine OCR errors while occasionally introducing new errors, particularly for words containing special characters. This outcome motivated our transition to a rule-based heuristic approach, which proved more reliable given the limited availability of Bahnar language resources. Figure \ref{fig:ocr_lm_pages56} illustrates representative language model outputs applied to pages 5 and 6, demonstrating the limitations of this approach.

\subsection{Processing Pipeline Architecture}
\begin{figure}[!htbp]
    \centering
    \includegraphics[width=\linewidth]{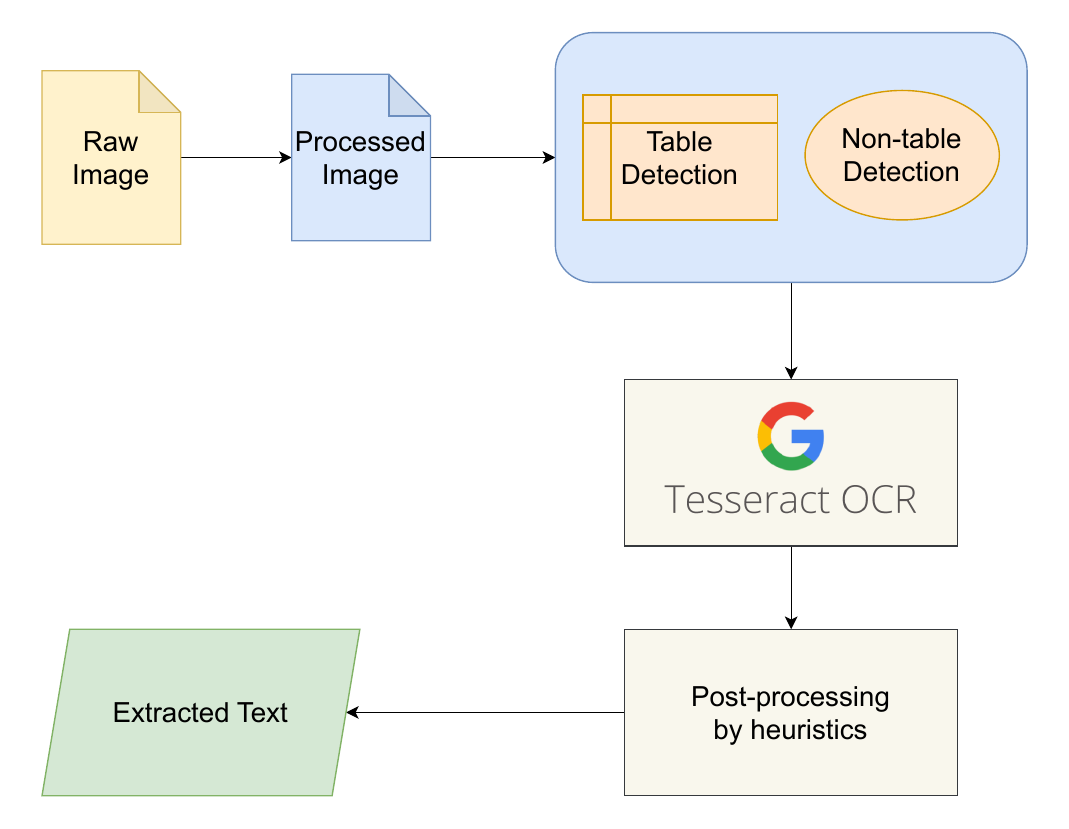}
    \caption{Architecture for Bahnar text extraction from images.}
    \label{fig:bahnar_pipeline}
\end{figure}

Figure \ref{fig:bahnar_pipeline} presents our end-to-end pipeline for Bahnar text extraction from digitized dictionary images. The pipeline consists of five sequential stages designed to handle the unique characteristics of the source material.

\subsubsection{Image Acquisition and Preparation}
PDF scans are converted to raster images and cropped to isolate regions of interest, removing margins and non-textual elements. This preprocessing reduces computational overhead and minimizes potential noise sources.

\subsubsection{Image Enhancement}
Raw images undergo preprocessing operations to enhance text-background contrast and remove noise artifacts. This stage applies adaptive thresholding and contrast adjustment techniques to improve character boundaries, which is critical for subsequent OCR accuracy.

\subsubsection{Structural Analysis}
Morphological operations (erosion and dilation) are applied to emphasize textual features, followed by Hough line detection to identify horizontal and vertical ruling lines. These detected lines enable precise segmentation of tabular structures and differentiation between table cells and peripheral content regions.

\subsubsection{Layout-Aware Text Extraction}
The pipeline bifurcates into parallel processing streams: one for tabular regions and another for non-tabular content. Tesseract OCR is applied to each region independently, preserving structural context for downstream formatting.

\subsubsection{Post-Processing and Formatting}
OCR outputs undergo heuristic-based correction to address systematic misrecognition of Bahnar-specific characters. Extracted text from table cells and non-tabular regions is then reformatted according to the detected table structure.

\section{Results}
\subsection{Baseline Performance Without Table Detection}
\begin{figure}[!htbp]
    \centering
    \begin{subfigure}[b]{\linewidth}
        \centering
        \includegraphics[width=\linewidth]{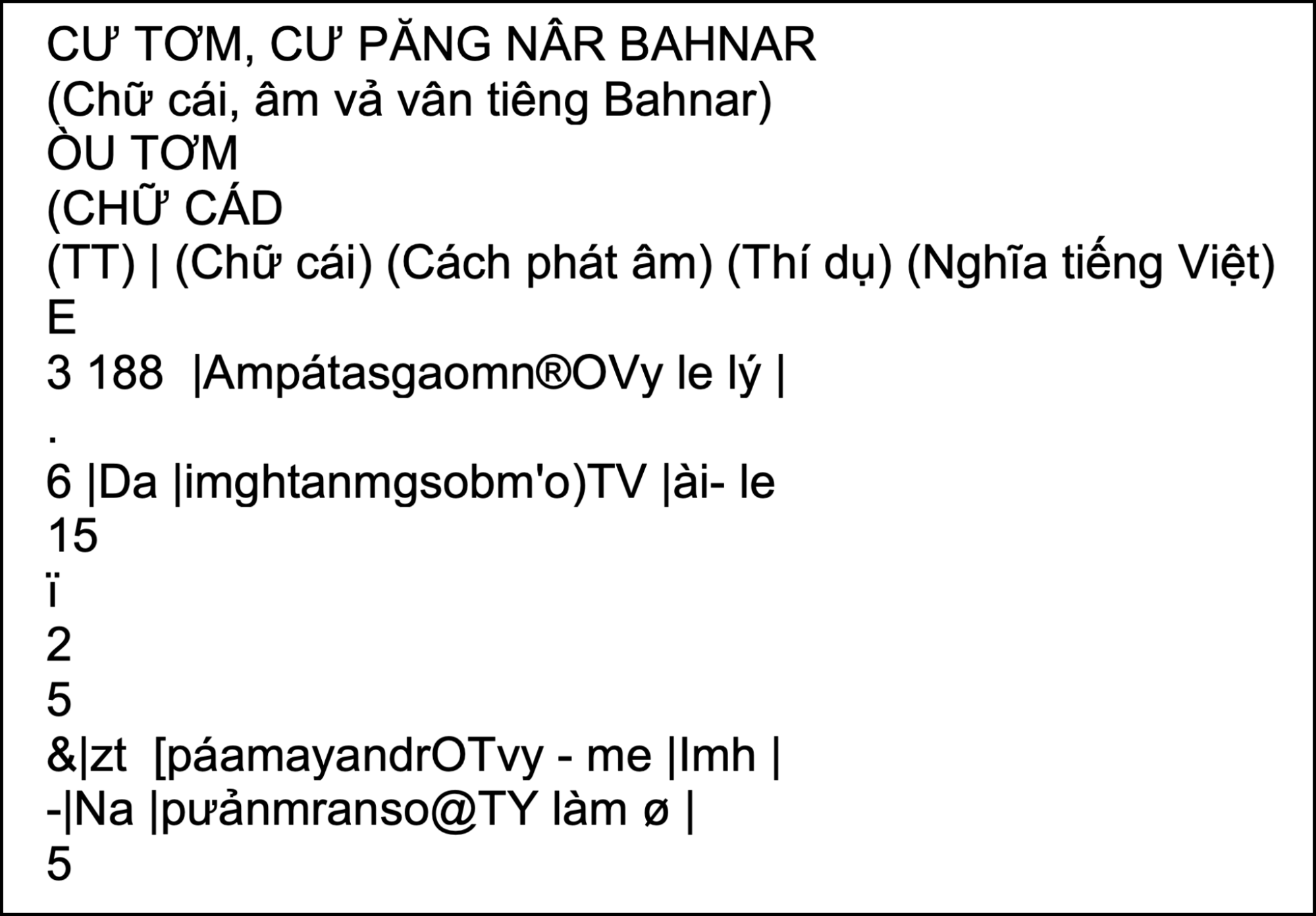}
    \end{subfigure}
    \begin{subfigure}[b]{\linewidth}
        \centering
        \includegraphics[width=\linewidth]{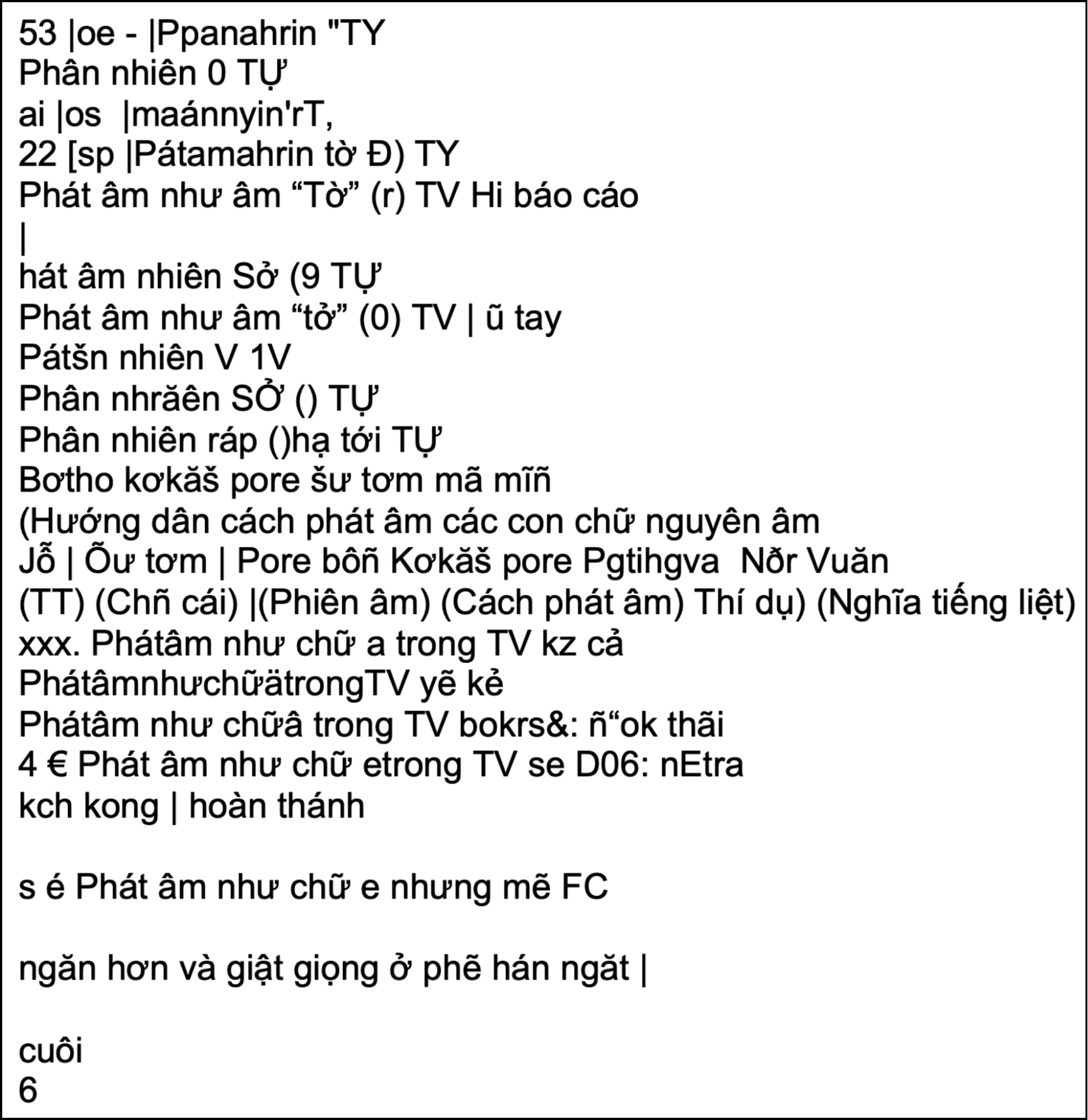}
    \end{subfigure}
    \caption{OCR output for pages 5 and 6 prior to table detection.}
    \label{fig:ocr_pre_table_pages56}
\end{figure}

Initial experiments were conducted without table detection to establish a baseline for comparison. Figure \ref{fig:ocr_pre_table_pages56} presents the OCR output for pages 5 and 6 under this configuration.

The baseline results reveal significant deficiencies when table detection is absent from the pipeline. Text extraction within tabular regions exhibits poor quality, with frequent misalignment and character recognition errors. The OCR engine fails to preserve the logical structure of the dictionary entries, resulting in outputs that are inconsistent with the source material. This degradation can be attributed to the lack of layout awareness, which causes the engine to process tabular and non-tabular content uniformly without accounting for their distinct structural properties.

\subsection{Performance With Table Detection}
\begin{figure}[!htbp]
    \centering
    \begin{subfigure}[b]{\linewidth}
        \centering
        \includegraphics[width=\linewidth]{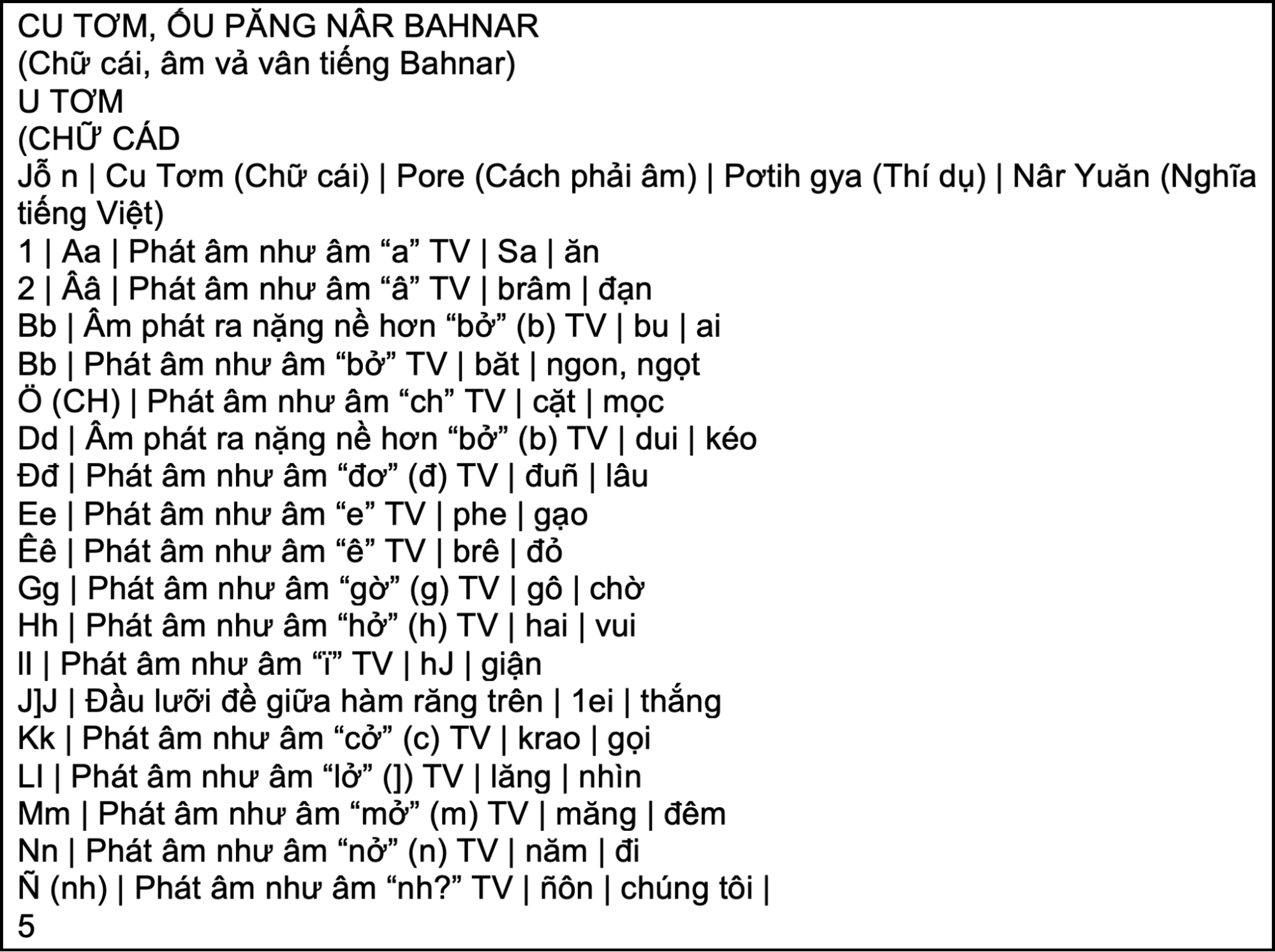}
    \end{subfigure}
    \begin{subfigure}[b]{\linewidth}
        \centering
        \includegraphics[width=\linewidth]{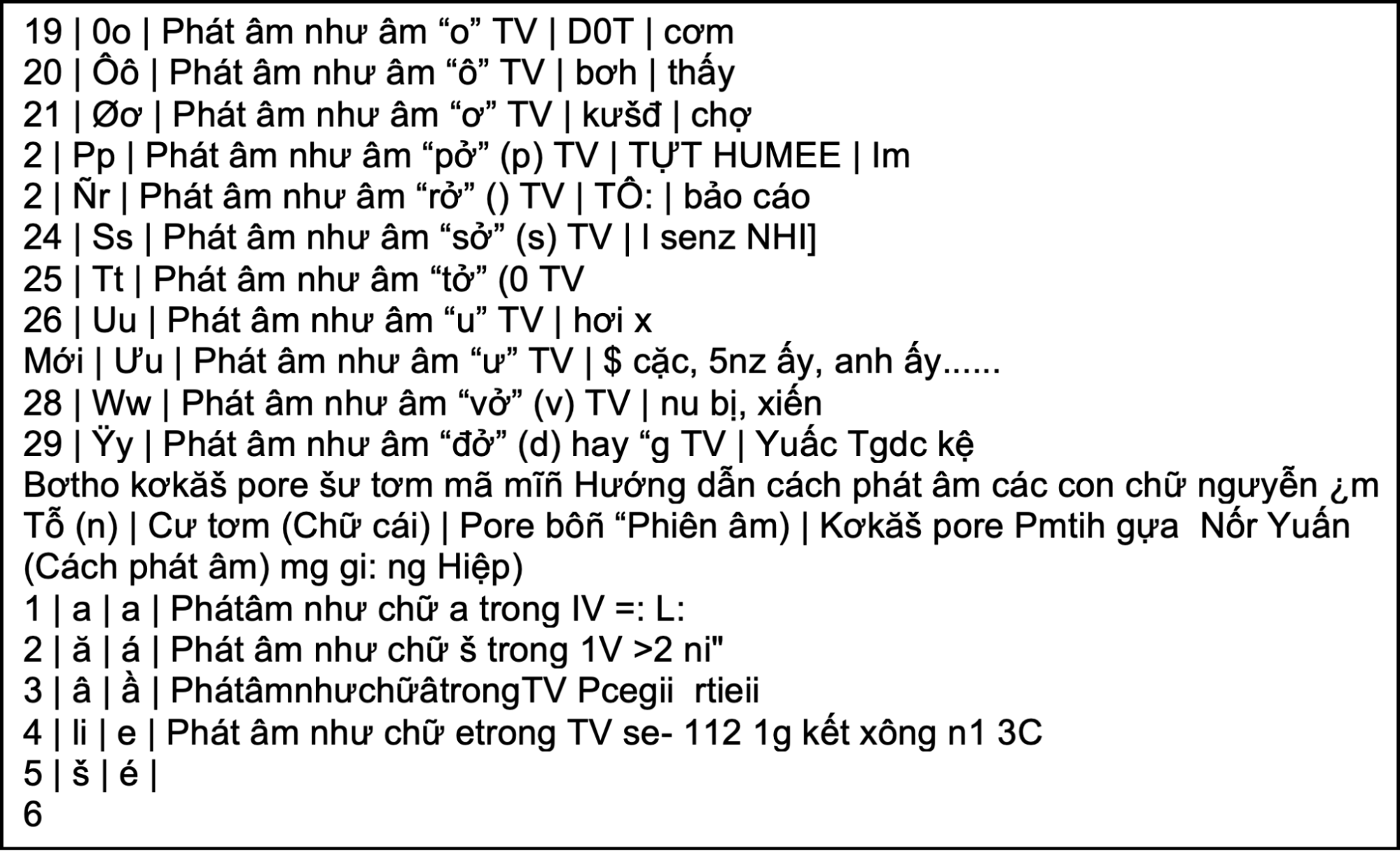}
    \end{subfigure}
    \caption{OCR output for pages 5 and 6 after table detection.}
    \label{fig:ocr_post_table_pages56}
\end{figure}

To address the limitations observed in the baseline configuration, we incorporated table detection into the pipeline. Figure \ref{fig:ocr_post_table_pages56} illustrates the OCR output for the same pages after implementing structural analysis and layout-aware text extraction.

Comparison between Figure \ref{fig:ocr_pre_table_pages56} and Figure \ref{fig:ocr_post_table_pages56} reveals clear improvements in extraction quality. The incorporation of table detection enables the pipeline to preserve the structural integrity of tabular content, maintaining proper alignment between columns and rows. Text extraction accuracy within table cells shows marked enhancement, with outputs exhibiting greater consistency with the source material. This improvement validates the efficacy of layout-aware processing for structured document content.

\subsection{Impact of Heuristic Post-Processing}
\begin{figure}[!htbp]
    \centering
    \begin{subfigure}[b]{\linewidth}
        \centering
        \includegraphics[width=\linewidth]{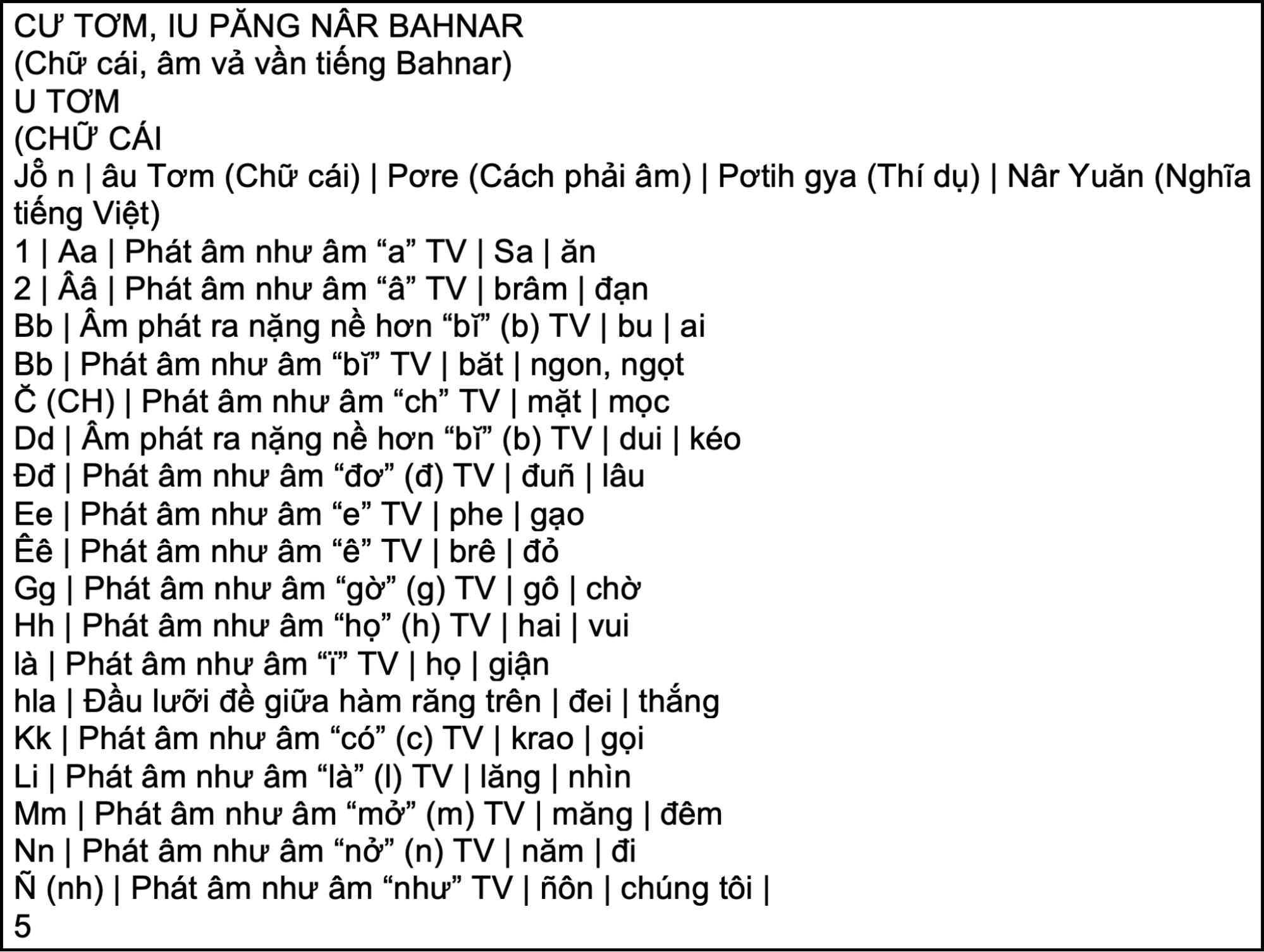}
    \end{subfigure}
    \begin{subfigure}[b]{\linewidth}
        \centering
        \includegraphics[width=\linewidth]{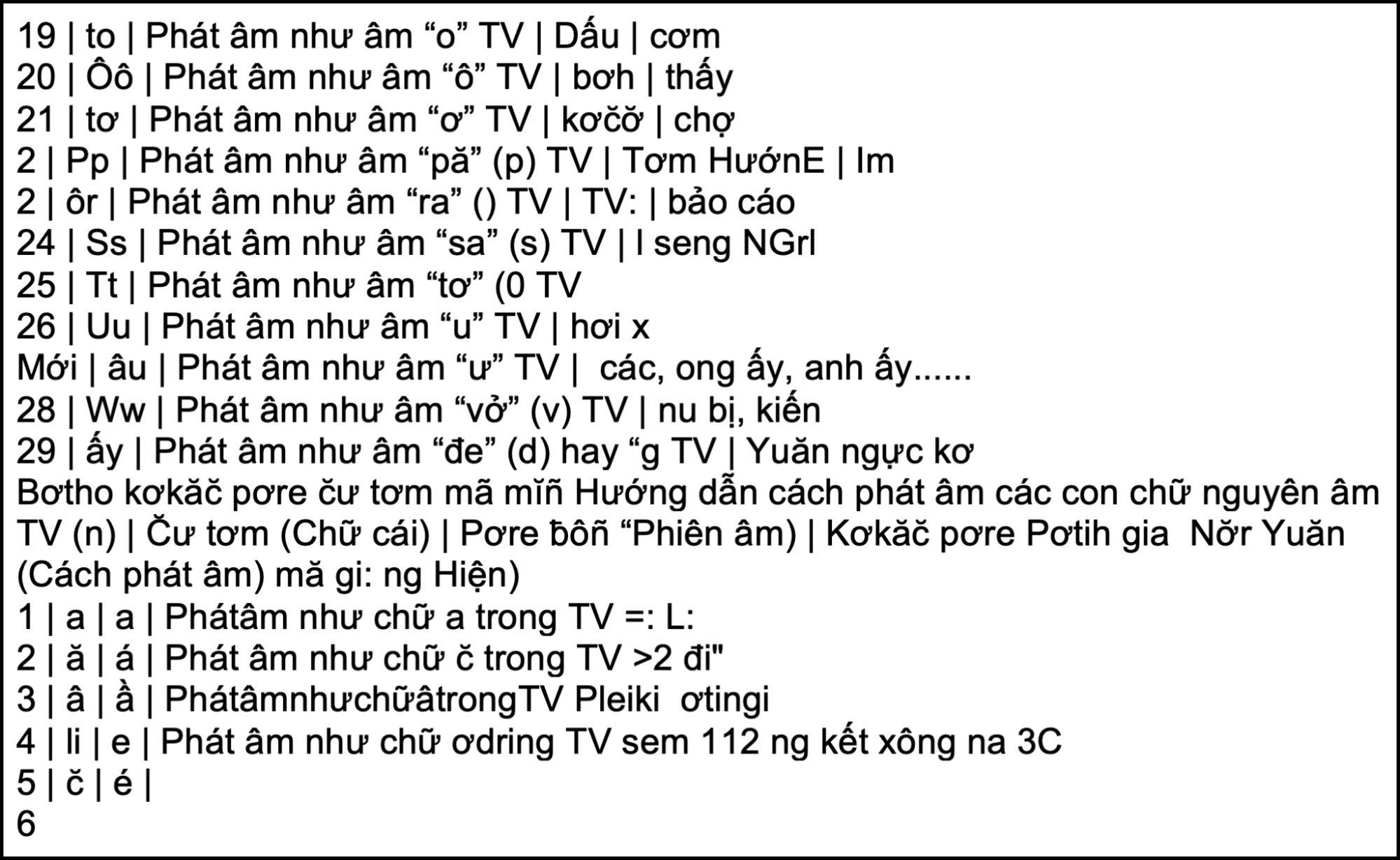}
    \end{subfigure}
    \caption{OCR output for pages 5 and 6 after heuristic post-processing.}
    \label{fig:ocr_post_heuristic_pages56}
\end{figure}

The final stage of our pipeline applies heuristic-based correction to address systematic character recognition errors specific to Bahnar orthography. Figure \ref{fig:ocr_post_heuristic_pages56} presents the output after applying these correction rules.

To quantitatively evaluate the effectiveness of heuristic post-processing, we randomly selected five pages from the dataset, comprising a total of 969 words. Comparison with ground truth showed that 706 words were correctly recognized before applying heuristics, while 768 words achieved correct recognition after post-processing. This represents an improvement of 62 words, corresponding to an accuracy increase from 72.86\% to 79.26\%.

\begin{table}[htbp]
    \centering
    \begin{tabular}{lcc}
        \toprule
        & \textbf{Before Heuristic} & \textbf{After Heuristic} \\
        \midrule
        Validation Accuracy & 0.7286 & 0.7926 \\
        \bottomrule
    \end{tabular}
    \caption{Validation accuracy before and after applying heuristic post-processing.}
    \label{tab:heuristic_accuracy}
\end{table}

Table \ref{tab:heuristic_accuracy} presents the quantitative validation results, indicating that heuristic post-processing yields a 6.4 percentage point improvement in accuracy. This substantial gain stems from the systematic correction of character substitution patterns that occur when Tesseract encounters Bahnar-specific diacritical marks absent from its training data.

Table \ref{tab:ocr_heuristic_comparison} provides representative examples of corrections achieved through heuristic post-processing. Common error patterns include the misrecognition of breve-marked characters (`\( \breve{\text{c}}\)', `\(\breve{\text{o}}\)', `\(\breve{\text{ơ}}\)', `\(\breve{\text{i}}\)', `\(\breve{\text{u}}\)') as alternative diacritical forms or similar-appearing characters. The heuristic rules successfully restore the correct orthographic representation by mapping these systematic errors to their intended characters based on contextual patterns and character frequency distributions in the Bahnar language.

\begin{table}[htbp]
    \centering
    \begin{tabular}{ccc}
        \toprule
        \textbf{Ground Truth} & \textbf{Before Heuristic} & \textbf{After Heuristic} \\
        \midrule
        kơkă$\breve{\text{c}}$  & kơkăš & kơkă$\breve{\text{c}}$ \\
        s$\breve{\text{o}}$k   & sốk   & s$\breve{\text{o}}$k   \\
        kơ$\breve{\text{o}}$$\breve{\text{ơ}}$   & kơšđ
       & kơ$\breve{\text{o}}$$\breve{\text{ơ}}$   \\
       {\bbar}ô\(\tilde{\text{n}}\)  & bô\(\tilde{\text{n}}\) & {\bbar}ô\(\tilde{\text{n}}\) \\
          ph$\breve{\text{ơ}}$k   & phỡk   & ph$\breve{\text{ơ}}$k   \\
           tơx$\breve{\text{i}}$   & tơxï   &  tơx$\breve{\text{i}}$   \\
           hơt$\breve{\text{u}}$t   & hơtũt   & hơt$\breve{\text{u}}$t   \\
           pơ\(\tilde{\text{n}}\)an   & po\(\tilde{\text{n}}\)an   & pơ\(\tilde{\text{n}}\)an   \\
           pơđôr   & pođØr   & pơđôr   \\
           N$\breve{\text{ơ}}$r   & Nốr   & N$\breve{\text{ơ}}$r   \\

        \bottomrule
    \end{tabular}
    \caption{Comparison of OCR output before and after heuristic post-processing.}
    \label{tab:ocr_heuristic_comparison}
\end{table}

The examples in Table \ref{tab:ocr_heuristic_comparison} illustrate the predominant error classes encountered during Bahnar text recognition. Characters such as `š', `ỡ', `ï', `ũ', and `Ø' represent systematic misrecognitions that occur when the OCR engine attempts to interpret Bahnar diacritics using its Vietnamese and English language models. The heuristic correction rules successfully resolve these substitutions, restoring the authentic Bahnar orthography and significantly improving the usability of the extracted text for subsequent linguistic analysis and lexicographic applications.

\section{Conclusion}
This work presents a comprehensive pipeline for Bahnar text extraction and recognition from digitized dictionary images. The presented method establishes the viability of adapting existing OCR technologies to low-resource minority languages through strategic integration of image preprocessing, structural analysis, layout-aware text extraction, and heuristic-based post-processing. The pipeline achieves a validation accuracy of 79.26\% on representative test samples, representing a 6.4\% gain over baseline configurations and validating the effectiveness of our methodology. 

This research contributes to the broader initiative of language preservation and digital documentation for ethnic minority communities in Vietnam. By successfully addressing the unique orthographic challenges posed by Bahnar special characters and diacritical marks, we establish a foundation for similar efforts targeting other Austroasiatic minority languages such as Rade and Sedang.

Future research directions include several promising avenues for enhancement. First, expanding the Bahnar lexical database will provide additional training data for refined heuristic rules and improved character recognition patterns. Second, we plan to investigate grammar-based correction methods as an alternative or complement to probabilistic approaches, potentially improving accuracy for context-dependent character disambiguation. Third, we aim to explore modern neural language models for syntactic-level processing of Bahnar text, enabling applications beyond simple character recognition such as grammatical analysis and semantic understanding. Finally, given the transferability demonstrated by our approach, we intend to extend this methodology to additional minority languages, contributing to the preservation and digitization of Vietnam's linguistic diversity.

\section*{Acknowledgment}
The authors thank the Ho Chi Minh City University of Technology and Vietnam National University Ho Chi Minh City for providing the necessary computational resources and facilities. We also gratefully acknowledge Nguyen Quang Duc for his insightful guidance throughout the research process.

\bibliographystyle{IEEEtran}
\bibliography{ref}

\end{document}

%% file: Tikz/spread_points.tex
\begin{tikzpicture}[
  x=1.5cm, y=1.5cm, 
  mynode/.style={circle, draw, blue!50, fill=white, minimum size=0.7cm},
  filled/.style={mynode, fill=teal!70}, 
    arrow/.style={->, thick, teal!80},
    bigarrow/.style={-{Implies}, line width=3pt, draw=gray!50}
]

\tikzset{
  pics/grid/.style = {
    code = {
      \colorlet{gridcolor}{black!70} 
      \foreach \x in {0,...,4} { \foreach \y in {0,...,2} { \pgfmathtruncatemacro{\ynext}{\y+1}; \draw[gridcolor] (\x, -\y) -- (\x, -\ynext); } }
      \foreach \y in {0,...,3} { \foreach \x in {0,...,3} { \pgfmathtruncatemacro{\xnext}{\x+1}; \draw[gridcolor] (\x, -\y) -- (\xnext, -\y); } }
      \foreach \x in {0,...,4} { \foreach \y in {0,...,3} { \node[mynode] (#1-\x-\y) at (\x, -\y) {}; } }
    }
  }
}

\begin{scope}
  \pic {grid=n1}; 
  \node[filled, label=above:{$(x_0, y_0)$}] at (0, 0) {};
\end{scope}

\begin{scope}[xshift=8cm]
  \pic {grid=n2}; 
  \node[filled, label=above:{$(x_0, y_0)$}] (n2-0-0) at (0, 0) {};
  \node[filled, label=above:{$(x_1, y_0)$}] (n2-1-0) at (1, 0) {};
  \node[filled, label=left:{$(x_0, y_1)$}]  (n2-0-1) at (0, -1) {};
  
  \draw[arrow] (n2-0-0) -- (n2-1-0);
  \draw[arrow] (n2-0-0) -- (n2-0-1);
\end{scope}

\begin{scope}[xshift=16cm]
  \pic {grid=n3};
  \node[filled, label=above:{$(x_0, y_0)$}] (n3-0-0) at (0, 0) {};
  \node[filled, label=above:{$(x_1, y_0)$}] (n3-1-0) at (1, 0) {};
  \node[filled, label=left:{$(x_0, y_1)$}]  (n3-0-1) at (0, -1) {};
  \node[filled, label={[below, xshift=20pt, yshift=-20pt]{$(x_1, y_1)$}}] (n3-1-1) at (1, -1) {};  
  \draw[arrow] (n3-1-0) -- (n3-1-1);
  \draw[arrow] (n3-0-1) -- (n3-1-1);
\end{scope}

\draw[bigarrow] (6.5cm, -2.25cm) -- (7.5cm, -2.25cm);
\draw[bigarrow] (14.5cm, -2.25cm) -- (15.5cm, -2.25cm);

\end{tikzpicture}